\documentclass{article}

\usepackage{arxiv}
\usepackage[hyphens]{url}  
\usepackage{graphicx} 
\usepackage{subfigure}
\usepackage{graphicx}  
\usepackage{amsmath}
\usepackage{amssymb}
\usepackage[utf8]{inputenc} 
\usepackage[T1]{fontenc}    
\usepackage{hyperref}       
\usepackage{url}            
\usepackage{booktabs}       
\usepackage{amsfonts}       
\usepackage{nicefrac}       
\usepackage{microtype}      
\usepackage{lipsum}
\usepackage{subfigure}
\usepackage{graphicx}  
\title{Region and Object based Panoptic Image Synthesis through Conditional GANs}

\author{
  Heng Wang\thanks{Authors contributed equally.} \\
  School of Computer Science\\
  The University of Sydney\\
  Sydney, Australia\\
  \texttt{hwan9147@uni.sydney.edu.au} \\
  \And
  Donghao Zhang\textsuperscript{*} \\
  School of Computer Science\\
  The University of Sydney\\
  Sydney, Australia\\
  \texttt{dzha9516@uni.sydney.edu.au} \\
  \And
  Yang Song \\
  School of Computer Science and Engineering\\
  University of New South Wales\\
  Sydney, Australia\\
  \texttt{yang.song1@unsw.edu.au} \\
  \And
  Heng Huang \\
  Department of Electrical and Computer Engineering\\
  University of Pittsburgh\\
  Pittsburgh, USA\\
  \texttt{henghuanghh@gmail.com} \\
  \And
  Mei Chen \\
  Microsoft Corporation\\ 
  Redmond, Washington, USA\\
  \texttt{may4mc@gmail.com} \\
  \And
  Weidong Cai \\
  School of Computer Science\\
  The University of Sydney\\
  Sydney, Australia\\
  \texttt{tom.cai@sydney.edu.au} \\
}

\begin{document}
\maketitle

\begin{abstract}
Image-to-image translation is significant to many computer vision and machine learning tasks such as image synthesis and video synthesis. It has primary applications in the graphics editing and animation industries. With the development of generative adversarial networks, a lot of attention has been drawn to image-to-image translation tasks. In this paper, we propose and investigate a novel task named as panoptic-level image-to-image translation and a naive baseline of solving this task. Panoptic-level image translation extends the current image translation task to two separate objectives of semantic style translation (adjust the style of objects to that of different domains) and instance transfiguration (swap between different types of objects). The proposed task generates an image from a complete and detailed panoptic perspective which can enrich the context of real-world vision synthesis. Our contribution consists of the proposal of a significant task worth investigating and a naive baseline of solving it. The proposed baseline consists of the multiple instances sequential translation and semantic-level translation with domain-invariant content code. 
\end{abstract}

\keywords{Image synthesis \and Panoptic image translation \and Generative adversarial networks}



\maketitle

\section{Introduction}
\label{sec:intro}
Image-to-image translation (IIT) is significant to many computer vision and machine learning tasks such as image and video synthesis~\cite{bansal2018recycle}. It plays an important role in applications for graphics editing~\cite{bau2019gandissect} such as Photoshop, and for animation~\cite{chen2018cartoongan}. In addition, many computer vision problems including image colorization~\cite{colorization}, outdoor scene editing~\cite{sceneedit}, semantic inpainting~\cite{inpainting}, and style transfer~\cite{styletransfer,styletransfer2} are essentially topic-specific image-to-image translation tasks. 

\begin{figure}[!t]
\centering
    \subfigure[Image style-level image-to-image translation.]{\includegraphics[width=0.4\textwidth]{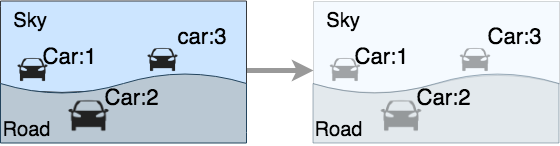}\label{fig:image_style}}
    \subfigure[Semantic-level image-to-image translation.]{\includegraphics[width=0.4\textwidth]{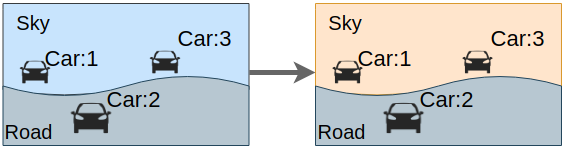}\label{fig:semantic}}
    \subfigure[Instance-level image-to-image translation.]{\includegraphics[width=0.405\textwidth]{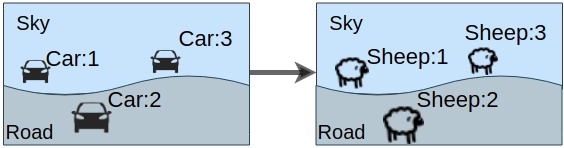}\label{fig:instance}}
    \subfigure[Panoptic-level image-to-image translation.]{\includegraphics[width=0.405\textwidth]{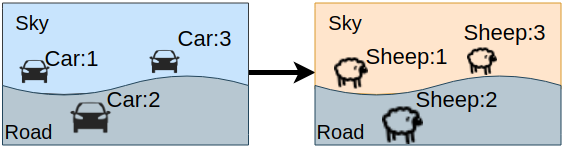}\label{fig:panoptic}}
    \caption{A schematic diagram of of different levels of image-to-image translation and the shade of each arrow indicates the difficulty of each image-to-image translation task.}
    \label{fig:intro}
\end{figure}

IIT can be divided into four different levels: (a) image style-level, (b) semantic-level, (c) instance-level, and (d) panoptic-level, as shown in Fig.~\ref{fig:intro}. The difficulty of IIT increases from (a) to (d). Image style-level translation transfers the overall style of an image to another domain. The simplest example is adjusting the contrast of an image. As shown in Fig.~\ref{fig:image_style}, the whole image including the semantic attributes (sky \& road) and instance attribute (car) have been lightened. The semantic-level translation shown in Fig.~\ref{fig:semantic} focuses more on semantic regions such as the sky and the road. The semantic-level IIT treats objects from the same category with the same label. Unlike the image style-level IIT, semantic-level IIT only converts the specified semantic attributes of an image from the source domain to another domain. In other words, the generated image preserves the style for most regions except those with the specified semantic attributes. For those semantic attributes to be translated, after translation, only the style has been changed while the semantic meaning is still preserved. As illustrated in Fig.~\ref{fig:semantic}, only the style of the sky has been changed while the style for the road and cars remains intact. Additionally, the instance-level IIT replaces the instance in the original image domain with another instance in the target domain. The translation could happen between two instances with different semantic meaning. We provide an example in Fig.~\ref{fig:instance}, instances of cars are expected to be transfigured to instances of sheep. In this paper, we format an interesting and challenging task: panoptic-level IIT. The panoptic image set consists of stuff-related semantic attributes such as sky and thing-related instance attributes like person and car. Current image translation tasks only consider image-level translation, i.e., the overall image style transfer, and instance-level translation, i.e., object transfiguration. As shown in Fig.~\ref{fig:panoptic}, our proposed task enables both these two translations. We also take the task a step further to ensure the translation for each panoptic-level attribute is independent, thus generating more creative images.

There are many applications where image-to-image translation (IIT) techniques can play an important role. Artist-level paintings generated by artificial intelligence instead of humans are impressive. Even though the translation results might be imperfect, some parts from the translated images are still useful given that the image style is another kind of art. Photo inpainting can also benefit from IIT techniques. For those valuable photos or paintings in the past, IIT techniques can be used to restore the original appearance of them. Apart from the significance of standard IIT techniques, panoptic-level IIT is also important in the following scenarios. The street scenes in real-time can be rendered to synthetic street scenes effortlessly when applying panoptic-level IIT. For instance, video games such as Grand Theft Auto would be the kind of application that can benefit from this panoptic-level IIT technique. Apart from electronic games, film productions can also benefit from these kinds of techniques. With the assistance of IIT, image editing tasks such as cartoonization can be more efficient. Furthermore, the animation industry could apply IIT techniques on raw materials directly to produce the synthesis effect rather than using time-consuming conservative post-production rendering. In the 2019 film Pokémon Detective Pikachu, there are lots of interactions between the animated character Pokémon Pikachu and real-world objects and environments. Assuming that panoptic-level IIT techniques are available, films like Pokémon Detective Pikachu can be recorded using real-world footage where a person would play the role of Pokémon Pikachu and later, during post-production, the person would be replaced directly by the character Pokémon Pikachu using panoptic IIT techniques. Similarly, documentary series recording footage of animals can be converted into other types of films, such as Pokémon fantasy films. In general, since any attributes could be translated independently, IIT techniques can be applied to the production of any urban fantasy applications.

To achieve the above applications, it is important to investigate panoptic-level IIT techniques. Even though image translation techniques have been developed in the past few years, panoptic-level IIT has not been investigated yet. In order to achieve panoptic-level IIT, one intuitive solution could be to implement instance-level IIT as the first step. However, translating instance-level objects requires datasets with at least instance-level annotations. From a different perspective, instance-level IIT can be decomposed into the combination of successful detection of each individual object, proper translation from the current object to a different object, and fine restoration of missing parts using algorithms such as semantic inpainting~\cite{inpainting}. The major issue is that translating the overall style such as color distribution might be easy, but the transfiguration of both position and shape of objects is challenging. For objects such as sheep and horses, they might be less difficult because they have similar structures. However, this is not the case for other objects such as sheep and cars since they are totally different. In terms of the semantic-level IIT, the main goal and challenge is to extract the difference while maintaining the same semantic meaning between different styles and to make the style transfer look natural. For example, the appearance of a road on a sunny day is different to the appearance of the same road on a foggy day. As an integration of these two tasks, panoptic-level IIT is no easier than either instance-level IIT or semantic-level IIT. The completion of panoptic-level IIT requires completing both the instance-level IIT and semantic-level IIT first. After that, ensuring the fusion of instance-level and semantic-level IIT is consistent is another interesting and challenging topic worth investigation.     


To solve the proposed panoptic-level IIT task, we design a systematic framework as baseline with modules to tackle translation in things and stuff respectively. To be more specific, the proposed framework includes a thing-related and a stuff-related attribute augmentation module. The transfiguration from cars into sheep is an example of the first thing-related attribute augmentation module. Once passing through the first module, the set of instance translated images are then converted by the stuff-related attribute augmentation module. During this process, style of specified stuff will be replaced with the style of region which has the same semantic meaning from another domain. We perform the evaluation on the COCO~\cite{lin2014microsoft}, CityScapes~\cite{cordts2016cityscapes}, and SYNTHIA~\cite{ros2016synthia} datasets. The COCO dataset was originally annotated for object detection, segmentation, and captioning. It holds different categories of things objects. The CityScapes and the SYNTHIA dataset provide real and synthetic images respectively with ground truth semantic segmentation. The evaluation results indicate that our proposed baseline achieves desired synthesized results for this novel task. In a nutshell, the major contributions of this paper can be concluded as: 
\begin{itemize}
\item We define and formulate a new task named as panoptic-level image-to-image translation. We hope the formulation of the panoptic-level task can inspire other researchers to investigate this problem in different ways and outperform our baseline.
\item We propose a simple but intuitive baseline of solving this task consisting of thing-related attribute augmentation and stuff-related attribute augmentation. 
\item We demonstrate the performance of the proposed baseline using the COCO, SYNTHIA, and CityScapes datasets.  
\end{itemize}

\section{Related Work}
\subsection{Image Style-Level Image-to-Image Translation}
Pix2Pix~\cite{pix2pix} is the first network to tackle the IIT with conditional GAN and L1 loss. However, Pix2Pix is only capable of generating low-resolution translated images. Pix2PixHD~\cite{pix2pixhd} improves Pix2Pix by introducing a multiscale discriminator, a robust loss function, and a more powerful generator. Nevertheless, previous work only focuses on the paired IIT. CycleGAN~\cite{cyclegan} solves the unpaired IIT by introducing the cycle-consistency loss. Similarly, the cycle-consistency loss is also applied in DualGAN~\cite{dualgan}. Network architectures for these two generators are identical. The generator is U-shaped and there are skip connections between the upsampling part and the downsampling part. To better learn the unsupervised IIT, UNIT~\cite{unit} makes a shared latent space assumption. It assumes that two different image domains have a common shared latent space. Given the shared latent-space code, images belonging to any domain can be recovered from it. The shared latent space assumption also implies cycle-consistency constraint proposed in previous methods. Unlike the assumption in UNIT, the  MUNIT~\cite{munit} assumes that latent space between different domains is partially shared. Before the MUNIT, IIT is one-to-one. To be more specific, the translation results are not diverse enough. The MUNIT solves the multimodal problem with within-domain reconstruction and cross-domain translation. In addition, IIT can also be tackled with disentangled representation~\cite{leedisentangle}.    
\begin{figure*}[!t]
\centering
\includegraphics[width=0.90\textwidth]{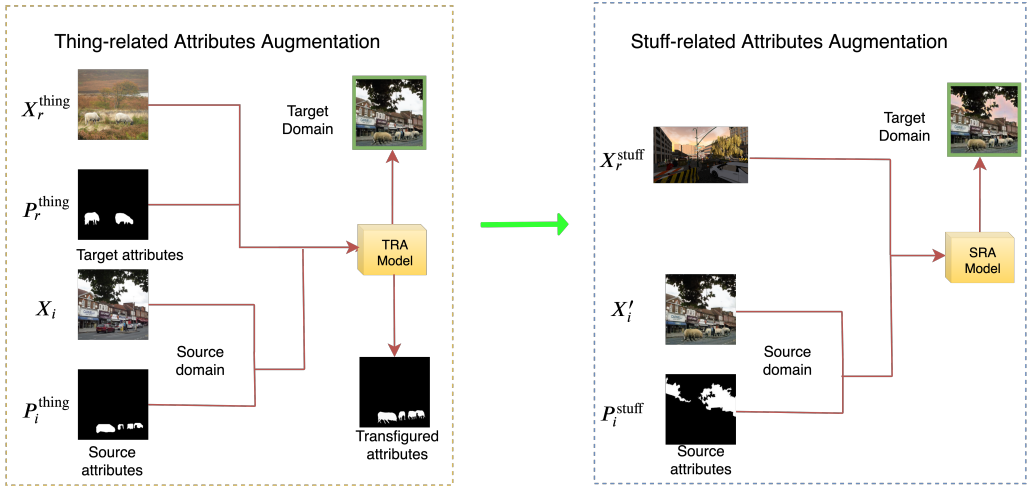} 
\caption{The overall view of the proposed panoptic-level image-to-image translation. $P_i^{\text{thing}}$ refers to the thing-related attributes from source domain $X_i$. $X_r^{\text{thing}}$ refers to the target domain containing the corresponding thing attributes $P_r^{\text{thing}}$ for $P_i^{\text{thing}}$ to translate. TRA and SRA stand for thing-related augmentation as described in Section~\ref{ssec:thing} and stuff-related augmentation as described in Section~\ref{ssec:stuff}. $X_i'$ refers to the intermediate translation domain after TRA. $X_r^{\text{stuff}}$ refers to the target domain for $X_i$ to conduct semantic-attributes augmentation on the source attributes $P_i^{\text{stuff}}$ from source domain $X_i$.}
\label{method_overall}
\end{figure*}
\begin{figure*}[!ht]
\centering
\includegraphics[width=0.90\textwidth]{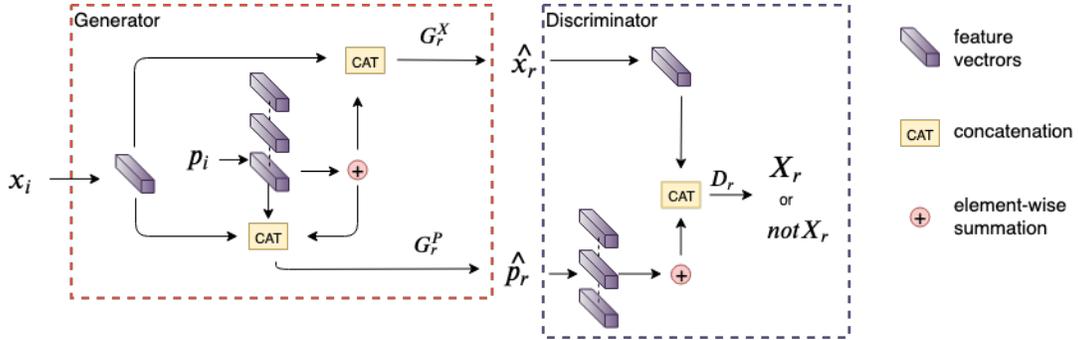} 
\caption{Training for thing-related attributes augmentation. $x_i$ refers to the input image whose thing-related attributes $p_i$ will be translated. $G_r^x$ and $G_r^p$ represent generators for the image and the thing-related attributes respectively. $\hat{x_r}$ and $\hat{p_r}$ refer to the generated image and the translated thing-related attributes respectively. $D_r$ represents the discriminator for image domain $X_r$.}
\label{method_instagan}
\end{figure*}
\begin{figure*}[!ht]
\centering
\includegraphics[width=0.9\textwidth]{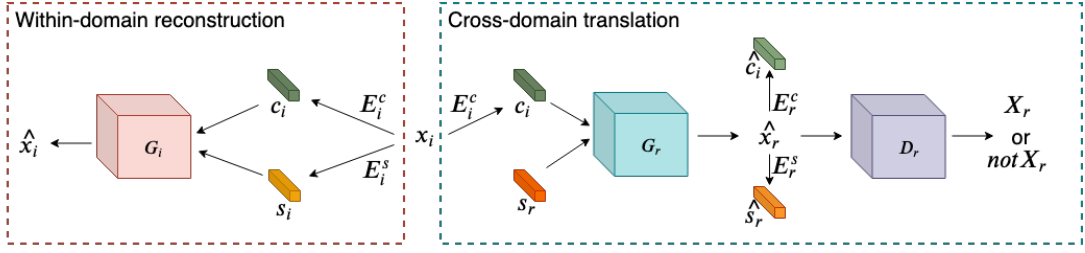} 
\caption{Training for stuff-related attributes augmentation. $x_i$ refers to the input image whose stuff-related attributes $p_i$ will be translated. $E_i^c$ and $E_i^s$ represent the content encoder and style encoder from domain $X_i$ respectively. $E_r^c$ and $E_r^s$ represent the content encoder and style encoder from domain $X_r$ respectively. $G_r$ and $G_i$ represent the generator for image domain $X_i$ and $X_r$ respectively. $D_r$ represents the discriminator for image domain $X_r$. $c_i$ and $\hat{c_i}$ refer to the content code produced by $E_i^c$ and $E_r^c$. $s_i$ refers to the style code produced by $E_i^s$. $s_r$ represents the random style sampled from domain $X_r$ while $\hat{s_r}$ refers to the style code produced by $E_r^s$.}
\label{method_muist}
\end{figure*}
\subsection{Instance-Level Image-to-Image Translation}
The instance-level IIT is more challenging than the image-style IIT. The image style-level IIT is more about preserving the image content and adjusting the style into a new image domain. The majority part of style translation is to adjust the image contrast. However, the instance-level IIT sometimes requires to the generation of parts in the target domain that do not exist in the source image. The instance-level image translation is a new research topic proposed by InstaGAN~\cite{instagan}. The InstaGAN applies the sequential mini-batch training mechanism. It reduces the memory constrains of the GPU when there are multiple instances. In terms of training loss of InstaGAN, there are multiple losses to improve the training including GAN loss for the domain loss, cycle-consistency loss, identity mapping loss, and context preseving loss. Furthermore, another work ~\cite{shen2019towards} defines the style code consisting of three parts: object style, background style, and global style. This work is based on the MUNIT framework~\cite{munit}, but improves MUNIT with instance-level GAN loss and instance-level reconstruction loss.

\subsection{Panoptic-Level Image-to-Image Translation}
To the best of our knowledge, we are the first to formulate the definition of the panoptic-level IIT task and propose a baseline aiming to inspire further research on this specific task. The panoptic-level IIT is inspired by the recent panoptic segmentation research work~\cite{ps}. In order to help readers gain a better understanding of the panoptic-level IIT, the methods of solving panoptic segmentation are reviewed here. The Cell R-CNN~\cite{zhang2018panoptic} attempts to solve this problem by unifying the semantic segmentation framework and instance segmentation framework~\cite{he2017mask}. The semantic branch of Cell R-CNN applies global convolution network (GCN) and the instance branch modifies the existing Mask R-CNN framework. The semantic branch and the instance branch share the same backbone. Recently, there have been many research works focusing on fusion of semantic segmentation and instance segmentation using the object-level and instance-level attentions~\cite{li2019attention}. UPSNet~\cite{xiong2019upsnet} integrates the semantic segmentation with instance segmentation by the panoptic head. The panoptic head is designed to merge the information from both semantic logits and mask logits. Instead of using original convolution, the semantic head of UPSNet applies the deformable convolution.       
\section{Methods}
\label{ssec:method}

\subsection{Task Format}
Given an image domain $X_i$, we translate $X_i$ to another image domain $X_i'$ by only augmenting its panoptic attributes $P_i$ to a set of different attributes $P_r$ originating from different image domains $X_r$. Formally, the translated image domain $X_i'$ can be formulated as:
\begin{equation}
    X_i' = \left\{X_i\backslash P_i\right\} \cap \left\{\cup_{P_r}\right\}
\end{equation}
where $\cup_{P_r}$ refers to the set of target attributes from different domains $X_r$. Then $X_i'$ represents the generated image domain with original attributes $P_i$ being replaced with $\cup_{P_r}$.

In detail, the set of panoptic attributes $\mathbf{p_i} \in P_i$ represents $M$ source elements to be translated in $X_i$: $\mathbf{p_i}$ = $\left\{p_{i}^{m}\right\}_{m = 1}^{M}$. These source elements are categorized based on panoptic segmentation of the given input image $x_i \in X_i$. That is, a selected source element can either be a thing (instance) or stuff (semantic only). On the other side, $\mathbf{p_r} \in \left\{\cup_{P_r}\right\}$ corresponds M target panoptic elements specified from different image domains: $\mathbf{p_r}$ = $\left\{p_{r}^{m}\right\}_{m = 1}^{M}$. $r$ refers to any image domains thus $p_{r}^{e}$ does not necessarily origin from the same image domain as $p_{r}^{v}$ for any $e$, $v \in [1, M]$. Then our proposed panoptic-level IIT task can be simply defined as:
\begin{equation}\label{task}
\begin{split}
X_i \times P_i \rightarrow  \left\{X_i\backslash P_i\right\} \times \left\{\cup_{P_r}\right\}\\
\end{split}
\end{equation}

As mentioned before, the panoptic attributes can be divided into two groups: 1) stuff-related semantic attribute set and 2) thing-related instance attribute set. Different translation rules should be applied on different panoptic attribute sets based on their properties. The translation rule regarding the two attribute sets are defined as follows:
\begin{itemize}
\item If the source element $p_i^m$ is considered as a semantic set, i.e., instances are ignored, the semantic content of $p_i^m$ should be preserved while only translating its style to $p_r^m$ of a target domain $X_r$. That is, we require $p_r^m$ to have the same semantic content as $p_i^m$. 
\item If instances are considered when translating $p_i^m$, i.e., $p_i^m$ is countable, $p_r^m$ can be instance attributes of different semantic meaning. Intuitively, the target element can have a different shape from the source element under this circumstance. That is, in addition to transferring style, we also aim to transfigure the shapes of instance objects. 
\end{itemize}
\subsection{Proposed Baseline}
When $M > 1$ and $\mathbf{p_i}$ combines both thing-related and stuff-related attributes, a simple system for the panoptic-level IIT task is to make reasonable heuristic to the output $\left\{x_i\backslash p_i^m\right\} \cap p_r^m$ for each $p_i^m \in \mathbf{p_i}$.

In detail, we group the $M$ source elements as $L$ stuff and $K$ things. We train $K$ thing-related attributes augmentation (TRA) models if there are $K$ thing-related source elements. We will discuss TRA model in Section~\ref{ssec:thing}. The first TRA model will be trained on the original data $x_i$. Before training the next TRA model for $p_i^{k+1} \rightarrow p_r^{k+1}$, we translate current updated image $x_{i, k}$ using current learnt $k$-th model and then pass the generated image $\left\{x_{i,k}\backslash p_i^k\right\} \times p_r^k$ as input data $x_{i,k+1}$ to next TRA model. The final translated image $x_i^{thing}$ after $K$ translations is $x_{i,K}$.

For $L$ stuff-related attributes $p_i^l$, if the corresponding $p_r^l$ are from $A$ different image domains, we will train $A$ different stuff-related attributes augmentation (SRA) models to learn all A image style-level mapping $X_i \rightarrow X_{r_{a=1}^{A}}$. SRA model will be elaborated in Section~\ref{ssec:stuff}. Unlike the sequential order when training $K$ TRA models, the training for $L$ SRA models can be conducted simultaneously with the same input $X_i$. 

To produce the final translated image, we apply the $L$ translated results of SRA models to $x_{i}^{thing}$ which can be reformulated as: 
\begin{equation}
\begin{split}
    & x_i'(p) = x_{i}^{thing}(p)\\
    & x_i'(p) = L \times \begin{cases}
    x_i'(p) & \text{if $B_i^l(p) = 0$}\\
    \hat{x_{r^l}}(p) & \text{otherwise}
  \end{cases}
\end{split}
\end{equation}
where $p$ represents the pixel location. $B_i^l(\cdot)$ refers to the $l$-th binary segmentation for the stuff-related attribute $p_i^l$ and $\hat{x_{r^l}}$ represents the $l$-th translated image from domain $X_i$ to domain $X_r$ using the corresponding SRA model. $L\times$ means to repeat the equation by $L$ times. The overall workflow is presented in Fig.~\ref{method_overall}. We firstly translate all the things-related-attributes using a set of TRA models. Then the partially translated image will be passed to the SRA models to get new semantic style for stuff-related attributes. This reasonable heuristic to the output of these two sets of attribute augmentation modules is simple but efficient as a baseline.

\subsection{Thing-related Attribute Augmentation}
\label{ssec:thing}
If the source scene element $p_i$ is countable, we can either translate $p_i$ to $p_r$ of the same semantic meaning following semantic style transfer as described in Section~\ref{ssec:stuff} or transfigure $p_i$ to another object $p_r$ which has different semantic content from another image domain $X_r$. It should be noted that $X_r$ could be the same as $X_i$. 

Assume that the target object $p_r$ appears in domain $X_r$ and $M$ is 1. We deploy InstaGAN~\cite{instagan} as our TRA model to learn the mapping $X_i \times P_i \rightarrow X_r \times P_r$. The InstaGAN model is trained from unpaired data $x_i \in X_i$ and $x_r \in X_r$. The thing-related set of attributes $\mathbf{p_i}$ is fed into the model sequentially along with the training data to incorporate the transfiguration information using the sequential mini-batch translation mechanism proposed in InstaGAN. Being consistent with InstaGAN, the thing-related attributes are also defined by the instance annotation masks. Hence, $p_i$ is a set of the instance segmentation of the thing-related attributes. As shown in Fig.~\ref{method_instagan}, the source image $x_i$ and its $p_i$ are encoded respectively to feature vectors.  We aim to translate each instance object to another instance object correspondingly. Thus, the encoded source image and the summation of encoded instance attributes will be incorporated together into the individual instance attribute to generate the new instance attribute $\hat{p_r}$ in domain $X_r$. Taking the feature vector of instance attributes into account, the generated $\hat{x_r}$ pays more attention to the instance objects. A discriminator $D_r$ is then deployed to tell whether the generated result is real enough to be considered as an image from domain $X_r$. Again, instance attributes are incorporated as additional information to the input of $D_r$. The process for $X_r\times P_r \rightarrow X_i\times P_i$ is similarly defined as Fig.~\ref{method_instagan}. Finally, the generated image is expected to preserve the information of the source image except for the target scene element specified by the thing-related attributes.

\subsection{Stuff-related Attribute Augmentation}
\label{ssec:stuff}
Many representations can be deployed to express the attributes. Intuitively, we use binary segmentation mask $B_i$ to identify the stuff-related semantic attributes. Suppose $M$ is 1. Given that the selected source element is uncountable, we require that the target source element $p_r^m$ should be of the same semantic meaning as $p_i^m$ but from another image domain $X_r$. We firstly learn the image style-level mapping between $X_i$ and $X_r$ by training the unsupervised translation model MUNIT~\cite{munit}. 

We also make the partially shared latent space assumption proposed in MUNIT to encourage non-deterministic mapping between $X_i$ and $X_r$. Sharing the same content space, the distribution $X_i$ and $X_r$ can be estimated using their style respectively. Consistent with MUNIT, we also assume the style code is sampled from the prior distribution $\mathcal{N}(0, \mathbf{I})$. Then, given a content code of $x_i$ and a random style code $s_r$, the generator for domain $X_r$ is trained to generate a synthesis image from domain $X_r$. A discriminator is required to guide the generative adversarial process. 

Following the partially shared latent space assumption proposed in MUNIT, the conditional probabilities $p(x_i\mid x_r)$ and $p(x_r\mid x_i)$ of unpaired data $x_i \in X_i$ and $x_r \in X_r$ are learnt by the adversarial loss and the joint reconstruction loss is defined as:
\begin{equation}
\begin{aligned}
    \min _{E_{i}, E_{r}, G_{i}, G_{r}} \max _{D_{i}, D_{r}} \mathcal{L}\left(E_{i}, E_{r}, G_{i}, G_{r}, D_{i}, D_{r}\right)=
    \mathcal{L}_{\mathrm{GAN}}^{x_{i}}+
    \mathcal{L}_{\mathrm{GAN}}^{x_{r}}+ \lambda_{x}\left(\mathcal{L}_{\text {recon }}^{x_{i}}+\mathcal{L}_{\text {recon }}^{x_{r}}\right)+\\
    \lambda_{c}\left(\mathcal{L}_{\text {recon }}^{c_{i}}+\mathcal{L}_{\text {recon }}^{c_{r}}\right)+
    \lambda_{s}\left(\mathcal{L}_{\text {recon }}^{s_{i}}+\mathcal{L}_{\text {recon }}^{s_{r}}\right)
    \end{aligned}
\end{equation}
where $\lambda_{x}$, $\lambda_{c}$, and $\lambda_{s}$ are hyper parameters to control the weights for each part of the loss. The two adversarial losses $\mathcal{L}_{\mathrm{GAN}}^{x_{i}}$ and $\mathcal{L}_{\mathrm{GAN}}^{x_{r}}$ are used to encourage the generated images to be indistinguishable from the respective target image domains. To learn the pair of encoder and decoder/generator, reconstruction loss is proposed in MUNIT. It includes the reconstruction of image $x_i \xrightarrow[]{\left(E_i^c, E_i^s \right)} c_i, s_i \xrightarrow[]{G_i(c_i, s_i)} \hat{x_i}$, content code $c_i \xrightarrow[]{G_r(c_i, s_r)} \hat{x_r} \xrightarrow[]{E_i^c} \hat{c_i}$, and style code $ s_r\xrightarrow[]{G_r(s_r, c_i)} \hat{x_r} \xrightarrow[]{E_i^s} \hat{s_r}$. The reconstruction for $x_r$, $c_r$, and $s_i$ as well as the process of $X_r \rightarrow X_i$ are defined similarly with the encoder $E_r: \left(E_r^s, E_r^c\right)$, generator $G_r$, and discriminator $D_i$. The training process of $X_i\rightarrow X_r$ is illustrated in Fig.~\ref{method_muist}. Encoder $E_i: \left(E_i^s, E_i^c\right)$ is used to project the input image $X_i$ to style $S_i$ and content $C_i$ latent space respectively. Then given a content code $c_i$ and a style code $s_r$, the generator $G_r$ will generate a translated image $\hat{x_r}$ which will be further examined by the discriminator $D_r$.

After training, we put the source image $x_i$ and random style code $s_r \sim \mathcal{N}(0, \mathbf{I})$ into the trained generator $G_r$ to generate the image $\hat{x_r}$ from the target domain $X_r$. We then incorporate the stuff-related attributes $B_i$ to $\hat{x_r}$ to get the panoptic translation $p_r$ for $p_i$. Note that the shape of $p_r$ is the same as $p_i$ while the intensity is augmented indicating the style transfer from $P_i$ to $P_r$. Finally, the panoptic-level translated image $x_r$ is defined as:
\begin{equation} \label{ssec:stuff_eq}
    x_r(p) = \begin{cases}
    x_i(p) & \text{if $B_i(p) = 0$} \\
    \hat{x_r}(p) & \text{otherwise}
  \end{cases}
\end{equation}
where $p$ represents the pixel location.
\begin{figure*}[!t]
\centering
\includegraphics[width=0.86\textwidth]{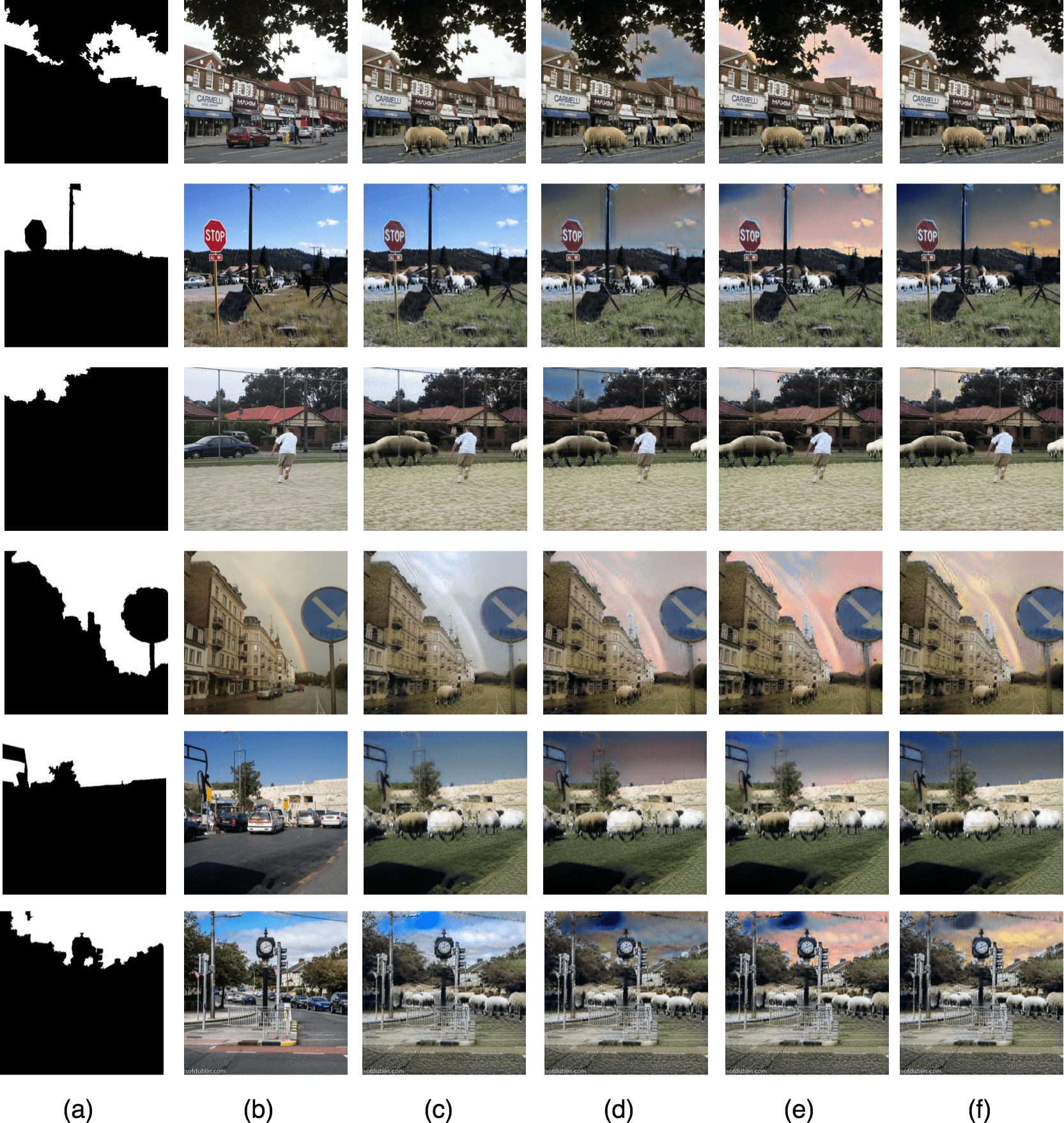} 
\caption{Panoptic-level image-to-image translation: (a) segmentation ground truth $B_i$ of stuff-related attributes for images from input domain $X_i$; (b) image from input $X_i$; (c) from car to sheep using the mapping $X_i \times P_i \rightarrow X_r \times P_r$; (d-f) the panoptic-level image-to-image translation results with different styles of sky region generated by the proposed baseline.}
\label{exp_pano}
\end{figure*}
\section{Experiment}
\subsection{Datasets and Implementation Details}
In this experiment, we translate two panoptic attributes from one domain to another two domains respectively to demonstrate our proposed baseline. These two panoptic attributes consist of one stuff-related attribute sky and one thing-related attribute car. Our baseline consists of a TRA model and a SRA model. The training for these two models are specified separately as follows.

The semantic style transfer translation is from a real sky scene to a synthetic one. We train such a translation using the real-world street scene dataset Cityscapes~\cite{cordts2016cityscapes} and the synthetic image dataset SYNTHIA~\cite{ros2016synthia}. Particularly, we use the SYNTHIA-RAND-CITYSCAPES subset which corresponds with the Cityscapes street scene dataset. We use all the 2975 images from the training set of Cityscapes as training images for one domain and 6196 random images from SYNTHIA-RAND-CITYSCAPES as training images for the other domain. The size of the Cityscapes images is $2048\times1024$ (height$\times$ weight) while that of the SYHTHIA images is $1280\times760$. We resize images from both datasets to $256\times256$ during the training of the SRA model for the translation $real \leftrightarrow synthetic$. This training process lasts 10800 iterations. We set hyperparameters $\lambda_{x}$, $\lambda_{c}$, and $\lambda_{s}$ as 10, 1, and 1 respectively.

To demonstrate the thing-related attributes augmentation, we decide to use images with cars and images with sheep from the dataset MS COCO~\cite{lin2014microsoft} since the two objects car and sheep vary in both shape and style. To train the TRA model, we use all the images containing the specified objects from 118287 COCO training images. That is, we use 10775 images as the car domain training set and 1516 images as the sheep domain training set. All images are resized to $220\times 220$ for efficient training. The training process contains 45 epochs.  

Both of these two models were trained from scratch using the optimizer Adam~\cite{adam} with batch size 1. We configure the training process on one GPU. We use 468 images with cars and 64 image with sheep from the validation set of MS COCO as testing images of our overall system. The stuff-related attribute sky is translated by 10 random styles learnt from the synthetic dataset while the thing-related attribute car is translated to sheep by applying the learnt generator and encoder.

\subsection{Results and Discussion}
We present some of our panoptic-level translation results in Fig.~\ref{exp_pano}. It consists of 6 images from 468 testing images with cars.  The attribute sky is indicated by the corresponding segmentation mask shown in column (a). Column (b) presents the original images from MS COCO validation images with cars. The results after thing-related attribute augmentation are displayed in column (c). We present 3 out of 10 random style translation results for the stuff-related augmentation in the last 3 columns (d), (e), and (f). 

As shown in column (b) and column (c), cars in the original images have been translated into sheep successfully. It is interesting to see that during the translation, feet of sheep grow where the tires of cars exist. The car to sheep translation should happen for every car instances in the source image. In the illustration, the number of cars in the image varies from 1 (the third row) to 9 (the last row). As clearly shown in the first image, each car has been translated to a sheep accordingly. However, under some circumstances, the translation might only be partially successful. Compared to the first image, in the fourth image, only the car in the front has been successfully translated into a sheep. The translation for cars near the back is blurred. The reason might be that their positions are far and the objects of interest turn out to be small-scale compared to the size of cars in the first image. It is challenging to preserve the outline of small-scale objects during IIT. 

Apart from the successful translation between two objects, the surrounding environment should be kept intact. As shown in the first image, the overall image has been preserved except for the car, for example, the white road lines still exist. However, there are minor artifacts on the surrounding environment. When we look at images from columns (b) and (c) carefully, even though cars have been successfully translated, the overall image is also affected. In particular, the area that near the objects of interest has suffered the most. For example, the road tends to be green. It is especially prominent in the last two images. This is due to the fact that sheep typically appears with grass. Hence, more constraints should be applied on keeping the neighbouring content intact. 

Since we use a random subset of SYNTHIA dataset, the training images are not consistent with seasons, weather, or illumination conditions. Therefore, there is no specific pattern in the translated animation-like style from column (d) to column (f). Nevertheless, our current focus is that we can translate the style of the stuff-related attributes. The visual effect of stuff-related attribute augmentation is inspiring. This is the first basic step.  

\section{Conclusion}
In this paper, we propose a novel task named panoptic-level image-to-image translation to translate any combination of a set of specific attributes in an image to any other image domains. Instead of changing the entire style to another domain like most current approaches, our proposed task expects panoptic-level translation which means the way each attribute is translated could be different and independent. The corresponding rules for translating different types of attributes are defined to make the task feasible and meaningful. For uncountable stuff attributes, it is expected to only translate the style while preserving the semantic meaning given that the shape of stuff attributes is consistent among different image domains. The semantic meaning of a stuff attribute is  unchanged. For countable thing attributes, we aim to transfigure the current object to another object with both style and shape being changed. Simple and efficient, our proposed system makes a heuristic combination of the outputs from two well-performing networks. Evaluated on several common datasets, our proposed framework achieves panoptic-level image-to-image translation in a consistent manner.

\bibliographystyle{unsrt}  
\bibliography{ref.bib}
\end{document}